\documentclass[letterpaper, 10 pt, conference]{ieeeconf}  

\IEEEoverridecommandlockouts                              

\usepackage{cite}
\usepackage{amsmath,amssymb,amsfonts}
\usepackage{algorithmic}
\usepackage{graphicx}
\usepackage{subfigure}
\usepackage{textcomp}
\usepackage{makecell}
\usepackage{color}

\title{\LARGE \bf
Toward Consistent Drift-free Visual Inertial Localization on Keyframe Based Map
}

\author{Zhuqing Zhang$^{1}$, Yanmei Jiao$^{1}$, Shoudong Huang$^{2}$, Yue Wang$^{1}$, Rong Xiong$^{1}$
\thanks{$^{1}$State Key Laboratory of Industrial Control Technology and Institute of Cyber-Systems and Control, Zhejiang University, Zhejiang, China. }
\thanks{$^{2}$Centre for Autonomous Systems, Faculty
of Engineering and Information Technology, University of Technology
Sydney, Ultimo, NSW 2007, Australia}
\thanks{Yue Wang is the corresponding author {\tt\small wangyue@iipc.zju.edu.cn}.}}

\begin{document}

\maketitle
\thispagestyle{empty}
\pagestyle{empty}

\begin{abstract}

Global localization is essential for robots to perform further tasks like navigation. In this paper, we propose a new framework to perform global localization based on a filter-based visual-inertial odometry framework MSCKF. To reduce the computation and memory consumption, we only maintain the keyframe poses of the map and employ Schmidt-EKF to update the state. This global localization framework is shown to be able to maintain the consistency of the state estimator. Furthermore, we introduce a re-linearization mechanism during the updating phase. This mechanism could ease the linearization error of observation function to make the state estimation more precise. The experiments show that this mechanism is crucial for large and challenging scenes. Simulations and experiments demonstrate the effectiveness and consistency of our global localization framework.

\end{abstract}

\section{Introduction and related works}

Autonomous mobile robot is a hot topic in contemporary research and development. To realize it, it is crucial to localize the robot in a global coordinate system. A good way to do that is employing simultaneous localization and mapping (SLAM). However, most of SLAM related works \cite{orbslam2,DSO,msckf,openvins,vinsmono} estimate robot state in a local coordinate system and the estimator would have drift. Global information could be introduced to solve this problem and therefore the robot system would be localized in a global frame. A commonly used global information for localization is global position system (GPS)\cite{vinsgps,gomsf,gpsvio}. However, GPS information is usually noisy and not always available. The pre-built visual map is another source of global information, which is what we consider in this paper. For visual-based localization system, the map usually contains keyframes' poses and related 2D or/and 3D features. When there are matches between the map and the online input images, the global information could be fused into the state estimator, and therefore remove the drift and increase the positional accuracy. 

There are lots of literatures performing global localization with pre-built visual maps \cite{vinsgps,getout,consistent,maplab,orbslam3}. However, how to use the information of the map is an open question. For a large scene, the map could contains hundreds of keyframes and tens of thousands of points, which is quite large in volume. Besides, the map information is actually not perfect and should have uncertainty information (covariance or information matrix), which makes the data much larger. To make the global localization system run in real-time, some trade-offs need to be made with map information.

A direct way of using a visual map is employing all the map information ($m$ keyframes' poses, $n$ landmarks, and their covariance or information matrices; $m<<n$ in general visual SLAM scenarios). However, since the computational complexity of updating covariance is $\mathcal{O}((m+n)^2)$, maintaining all the map information in the state vector would be computational expensive. To solve this problem, \cite{consistent} proposes a system based on Schmidt-EKF \cite{schmidtekf}. With the help of Schmidt-EKF, updating covariance would be less expensive ($\mathcal{O}(n)$). However, \cite{consistent} still stores the whole map in the state vector, which makes the state dimension quite high and needs a large memory consumption ($\mathcal{O}(n^2)$) to record information matrices. There are also some works \cite{vinsgps,maplab,orbslam3,tcgp} model maps as a set of keyframes. Maplab \cite{maplab} employs the ROVIO \cite{rovio1,rovio2} to realize local positioning, and combines the technique in \cite{getout} to obtain the global state. Its map is a set of pose-graph consisting of keyframes and landmarks (memory consumption $\mathcal{O}(m+n)$). In this framework, to improve computational efficiency, it does not maintain the map information in the state (computational complexity is constant), so that the correlation between current state and map information cannot be maintained, and therefore, the system would suffer from inconsistency. Besides, this system ignores the uncertainty of the map landmarks, and utilizes them in the observation function as fixed values, which again makes the system inconsistent. VINS-fusion \cite{vinsmono,vinsgps,vinslocal,vinscalib} builds the map in a similar way to Maplab\cite{maplab}, however it omits the landmarks information (memory consumption $\mathcal{O}(m)$). Its global observation function is constructed by current local 3D landmarks and matched 2D features in the map. Nevertheless, this framework cannot take the covariance of the map information into consideration, and results in inconsistency. Besides, in VINS-Fusion, local 3D landmarks are estimated online from the extracted 2D features in query images (not from the pre-built map) , which is coupled with the local odometry (or state estimator). To guarantee the state estimator runs in real time, the algorithm for extracting 2D keypoints and computing descriptors needs to be lightweight (e.g. FAST\cite{fast} + Brief\cite{brief}). These kinds of features, however, are not robust for matching with map features. Furthermore, the nonlinear optimization strategy employed in this framework is also time-consuming.
      
In this paper, we propose a filter-based global localization framework with consideration of the balance between accuracy, consistency, computation speed and memory space. To be specific, the pre-built map is made up of $m$ keyframes (pose and covariance) and $n$ landmarks (position). As $m$ is much smaller than $n$, the storage of this scheme is acceptable ($\mathcal{O}(m^2+n)$). Then, we maintain the poses of the map keyframes in the state vector, so that the stored covariance could be taken into consideration to make the system consistent. To avoid the huge computation of updating the whole map information, Schmidt-EKF is utilized to make computational complexity $\mathcal{O}(m)$. Furthermore, we also deal with a common problem in large-scale long-term navigation that the same scene in current images and the map images differs greatly: Our global observation function is based on the matching between 3D landmarks in the pre-built map and 2D features in the current query image, which is different from VINS-Fusion. In this way, matching procedure is decoupled from the online feature tracking, so that the real-time requirement is reduced and more robust features like R2D2\cite{r2d2} could be utilized to match the map and the query images. Therefore, the ability of long-term localization could be improved dramatically. Moreover, to avoid regarding landmarks as constants, the landmarks' positions from the map are used as linearization points of the observation function and the left null space projection is employed to handle the related Jacobians. Finally, in the case of the long time lack of global matching information, map landmarks are also used in our proposed re-linearization mechanism to improve the quality of the linearization points and therefore improve the accuracy of the localization. We call our method as consistent Schmidt-MSCKF localization (CS-MSCKF).






\section{System State and Observation Function}
\label{system}
\begin{figure*}[!t]
\vspace{3mm}
    \centering
    \setlength{\abovecaptionskip}{0cm}
    \includegraphics[width=0.98\textwidth]{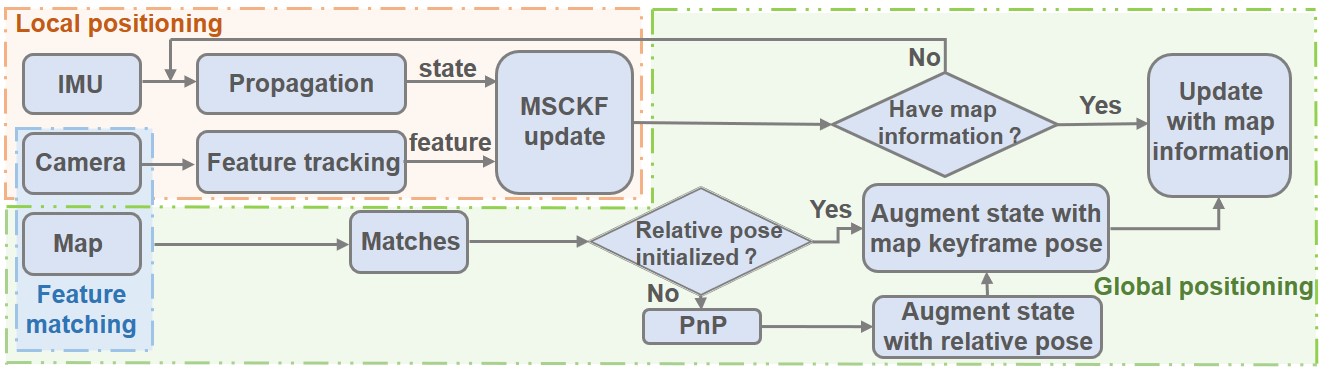}
     \caption{The overview of proposed global localization framework}
     \label{fig1}
     \vspace{-0.6cm}
\end{figure*}

As shown in Fig. \ref{fig1}, the whole system consists of three parts, feature matching with a given map, local positioning and global positioning.

\subsection{Local Positioning Based on MSCKF}
Local positioning in our work is actually a standard MSCKF\cite{msckf}, and this part follows the implementation of Open-VINS \cite{openvins}. As is shown in Fig. \ref{fig2}, local positioning is estimating the current state of the robot in local reference system {L}. Without considering the map information, we maintain the following state vector at time step $k$:

\begin{equation}
    \mathbf{x}_k= \left[\begin{array}{cc}
        \mathbf{x}_I^{\top} & \mathbf{x}_C^{\top}  \\ \end{array} \right]^{\top},
\end{equation}
where $\mathbf{x}_I$ is the current state of the robot. $I$ refers to the IMU reference. Here, we assume the body reference system is identical to the IMU reference system. We define $\mathbf{x}_I$ as 
\begin{equation} \label{eq:imu_state}
    \mathbf{x}_I=\left[\begin{array}{ccccc} {}^{I_k}\mathbf{q}_{L}^{\top}& {}^{L}\mathbf{p}_{I_k}^{\top} & {}^{L}\mathbf{v}_{I_k}^{\top} & \mathbf{b}_g^{\top}& \mathbf{b}_a^{\top}\\ \end{array} \right]^{\top},
\end{equation}
where ${}^{I_k}\mathbf{q}_{L}$ is an unit quaternion variable, defined by JPL\cite{jpl}. Its corresponding rotation matrix ${}^{I_k}\mathbf{R}_L$ rotates a 3D vector from the local  coordinate system ${L}$ into the IMU coordinate system ${I}$. ${}^{L}\mathbf{p}_{I_k}$ is the position of the body in the local coordinate system at time step $k$. ${}^{L}\mathbf{v}_{I_k}$ is the velocity of the body in the local coordinate system. $\mathbf{b}_g$ and $\mathbf{b}_a$ are the gyroscope and accelerometer bias. 
\begin{equation}
    \mathbf{x}_C= \left[\begin{matrix} {}^{I_{k-1}}\mathbf{q}_{L}^{\top}& {}^{L}\mathbf{p}_{I_{k-1}}^{\top}&\dots &{}^{I_{k-N}}\mathbf{q}_{L}^{\top}& {}^{L}\mathbf{p}_{I_{k-N}}^{\top}\\ \end{matrix} \right],
\end{equation}
which consists of the clones of the $N$ newest historical poses. $\mathbf{x}_C$ is the so-called sliding window.

\subsubsection{State Propagation}\label{prop}
When we receive IMU data at time step $k$, we could employ the IMU kinematics equations to predict the body state at time step $k+1$:
\begin{equation}\label{eq:state_prop}
    \mathbf{x}_{k+1|k}=\mathbf{f}(\mathbf{x}_{k},\mathbf{a}_{m_k}-\mathbf{n}_{a_k},\mathbf{\omega}_{m_k}-\mathbf{n}_{\omega_k}),
\end{equation}
where $\mathbf{f}$ represents the IMU kinematics equations, $\mathbf{a}_{m_k}$ and $\mathbf{\omega}_{m_k}$ are the observations (linear acceleration and angular velocity) from the IMU. $\mathbf{n}_{a_k}$ and $\mathbf{n}_{\omega_k}$ are the corresponding zero-mean Gaussian white noise of the IMU observations. With this function, we could propagate the state covariance by
\begin{equation}\label{eq:prop}
    \mathbf{P}_{k+1|k}=\mathbf{\Phi}_{k}\mathbf{P}_{k|k}\mathbf{\Phi}_{k}^{\top}+\mathbf{G}_{k}\mathbf{Q}_{k|k}\mathbf{G}_{k}^{\top},
\end{equation}
where $\mathbf{P}_{k|k}$ and $\mathbf{Q}_{k|k}$ are the system state covariance and the noise covariance respectively. $\mathbf{\Phi}_{k}$ and $\mathbf{G}_{k}$ are the Jacobians of (\ref{eq:prop}) with respect to the system state and the noise respectively. For detailed derivation, please refer to \cite{msckf, openvins}. 

\subsubsection{Observation Function}
In the framework of MSCKF, before fusing feature observations by the observation function, we could get a list of features $\left\{f \right\}$ and related cloned poses $\left\{\mathbf{T}_C\right\}$. For each feature $f_i$ observed by $\mathbf{T}_{C_j}$, we have the following observation function:
\begin{equation}
    \mathbf{z}_{f_i}=\mathbf{h}\left(\mathbf{T}_{C_j},\mathbf{f}_i\right)+ \mathbf{n}_{f_i}, 
\end{equation}
where $\mathbf{z}_{f_i}$ is the 2D observation of $f_i$ in the image. $\mathbf{f}_i$ is the 3D coordinate of $f_i$. By applying first-order Taylor expansion to the above observation function at the current estimated state, we could get the following functions:
\begin{equation}\label{eq:local_ob}
    \mathbf{z}_{f_i} \approx \mathbf{h}\left(\hat{\mathbf{x}}_k, \hat{\mathbf{f}}_i \right) + \mathbf{H}_{\hat{\mathbf{x}}_k}\tilde{\mathbf{x}}_k+\mathbf{H}_{\hat{\mathbf{f}}_i}\tilde{\mathbf{f}}_i+\mathbf{n}_{f_i},
\end{equation}
\vspace{-0.6cm}
\begin{equation}
    \mathbf{r}_{f_i}:=\tilde{\mathbf{z}}_{f_i}= \mathbf{H}_{\hat{\mathbf{x}}_k}\tilde{\mathbf{x}}_k+\mathbf{H}_{\hat{\mathbf{f}}_i}\tilde{\mathbf{f}}_i+\mathbf{n}_{f_i},
\end{equation}
where $\hat{\cdot}$ represents the estimated values and $\tilde{\cdot}$ represents the errors between true values and estimated values.  $\mathbf{H}_{\hat{\mathbf{x}}_k}$ and $\mathbf{H}_{\hat{\mathbf{f}}_i}$ are the Jacobians of $\mathbf{h}$ with respect to the estimated state and the observed feature respectively. As we do not maintain features in the state vector, $\tilde{\mathbf{f}}_i$ is marginalized by projecting $\mathbf{r}_{f_i}$ into the left null space of $\mathbf{H}_{\hat{\mathbf{f}}_i}$, $\mathbf{N}$:
\begin{equation}\label{eq:local_ns}
    \mathbf{N}^{\top} \mathbf{r}_{f_i}=\mathbf{N}^{\top}
\mathbf{H}_{\hat{\mathbf{x}}_k}\tilde{\mathbf{x}}_k+\mathbf{N}^{\top}\mathbf{H}_{\hat{\mathbf{f}}_i}\tilde{\mathbf{f}}_i+\mathbf{N}^{\top}\mathbf{n}_{f_i},
\end{equation}
\vspace{-0.6cm}
\begin{equation}\label{eq:local_error}
    \mathbf{r}^{\prime}_{f_i}=
\mathbf{H}^{\prime}_{\hat{\mathbf{x}}_k}\tilde{\mathbf{x}}_k+\mathbf{n}^{\prime}_{f_i},
\end{equation}
where $\mathbf{r}^{\prime}_{f_i}= \mathbf{N}^{\top} \mathbf{r}_{f_i}$, $\mathbf{H}^{\prime}_{\hat{\mathbf{x}}_k}=\mathbf{N}^{\top}
\mathbf{H}_{\hat{\mathbf{x}}_k}$, $\mathbf{N}^{\top}\mathbf{n}_{f_i}=\mathbf{N}^{\top}\mathbf{n}_{f_i}$. We would employ $\mathbf{H}^{\prime}_{\hat{\mathbf{x}}_k}$ and $\mathbf{r}^{\prime}_{f_i}$ to do the state update.

\subsection{Consistent Global Positioning}
\label{global model}


\begin{figure}[t!]
    \centering
    \setlength{\abovecaptionskip}{0cm}
    \includegraphics[width=0.8\linewidth]{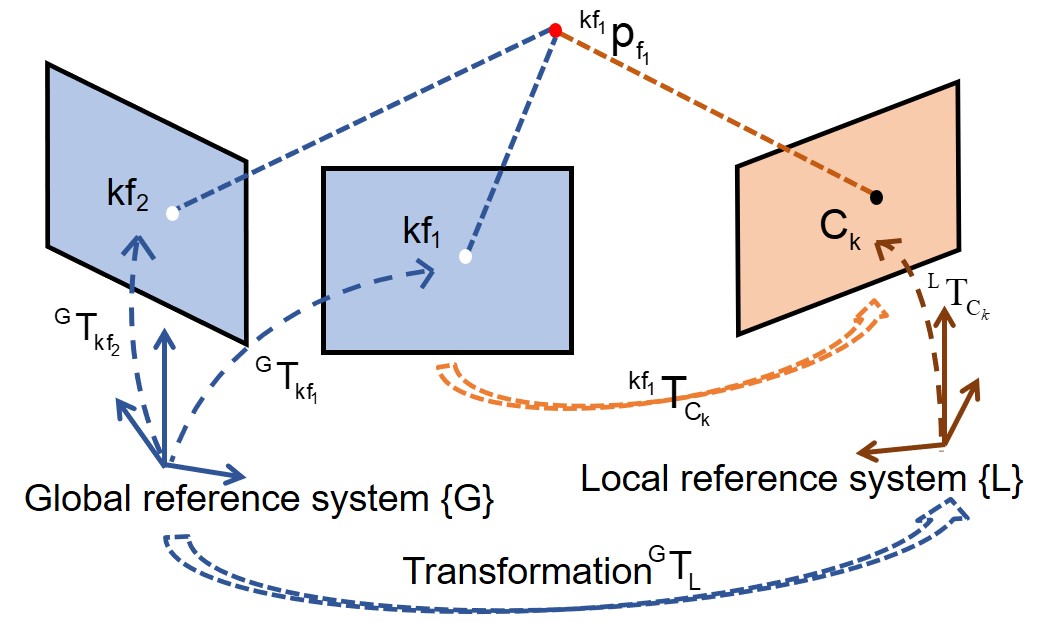}
     \caption{The global observation function}
     \label{fig2}
     \vspace{-0.6cm}
\end{figure}

\subsubsection{Initialization}
When there is a match between the current image ($\mathbf{C}_k$ in Fig. \ref{fig2}) and a keyframe in the pre-built map ($\mathbf{kf_1}$ in Fig. \ref{fig2}), we could get many corresponding 3D-2D feature matches (the red dot and the black dot in Fig. \ref{fig2}). We represent the 3D features in the coordinate system of the matched keyframe (anchored keyframe), and through EPnP\cite{epnp} we could get the relative pose ${}^{\mathbf{kf_1}}\mathbf{T}_{\mathbf{C}_k}$ between the keyframe and the current frame. This value, combined with the matched keyframe pose ${}^{G}\mathbf{T}_{\mathbf{kf_1}}$ and the current frame pose ${}^{L}\mathbf{T}_{\mathbf{C}_k}$, would be used to initialize ${}^{G}\mathbf{T}_{L}$. After initialization, ${}^{G}\mathbf{T}_{L} \triangleq \mathbf{x}_t $ would be added into the state vector:
\begin{equation}
    \mathbf{x}_k= \left[\begin{array}{ccc}
        \mathbf{x}_I^{\top} & \mathbf{x}_C^{\top} & \mathbf{x}_t^{\top}  \\ \end{array} \right]^{\top}.
\end{equation}
Besides, the covariance of the state would also be augmented. As the initial estimation of $\mathbf{x}_t$ may be inaccurate, its covariance could be assigned with a big value. It should be noted that every time we have matches from the map, the related keyframes' poses should be added into the state vector:
\begin{equation}
    \mathbf{x}_k= \left[\begin{array}{cccc}
        \mathbf{x}_I^{\top} & \mathbf{x}_C^{\top} & \mathbf{x}_t^{\top} & \mathbf{x}_{KF}^{\top} \\ \end{array} \right]^{\top},
\end{equation}
\begin{equation}
    \mathbf{x}_{KF}= \left[\begin{matrix}
        {}^{G}\mathbf{q}_{kf_1}^{\top}& {}^{G}\mathbf{p}_{kf_1}^{\top}&\dots &{}^{G}\mathbf{q}_{kf_M}^{\top}& {}^{G}\mathbf{p}_{kf_M}^{\top} \\ \end{matrix} \right]^{\top}.
\end{equation}
Also, the covariance of keyframes needs to be considered.
For the sake of simplicity, we divide the state into two parts: the active part $\mathbf{x}_A=\left[\begin{matrix}
        \mathbf{x}_I^{\top} & \mathbf{x}_C^{\top} & \mathbf{x}_t^{\top}\\ \end{matrix} \right]^{\top}$, and the nuisance part $\mathbf{x}_N=\mathbf{x}_{KF}$, so that
\begin{equation}\label{eq:xaxn}
    \mathbf{x}_k= \left[\begin{array}{cc}
        \mathbf{x}^{\top}_{A_k} & \mathbf{x}_{N_k}^{\top}\\ \end{array} \right]^{\top}.
\end{equation}
This kind of partition is very useful for Schmidt-EKF to be introduced in Sec. \ref{schmidt ekf}.

\subsubsection{Observation Function}
As is shown in Fig. \ref{fig2}, supposing the current image matches with multiple frames (for example, $\mathbf{kf_1}$ and $\mathbf{kf_2}$), for each 3D landmark (for example, $^{\mathbf{kf}_1}\mathbf{p}_{f_1}$), we could project it from the anchored keyframe ($\mathbf{kf}_1$) coordinate system into the 2D current image coordinate system by a nonlinear observation function $^{1}\mathbf{g}$:
\begin{equation}\label{eq:global_observation_function}
    {}^{1}\mathbf{z}_{f}={}^{1}\mathbf{g}(^{G}\mathbf{T}_{\mathbf{kf}_1},^{G}\mathbf{T}_{L},^{L}\mathbf{T}_{\mathbf{C}_k},^{\mathbf{kf}_1}\mathbf{p}_{f_1})+ {}^{1}\mathbf{n}_f,
\end{equation}
where ${}^{1}\mathbf{z}_{f}$ is the 2D pixel coordinate in the current image. Linearizing above function at the estimated values, just like (\ref{eq:local_ob}), we could get
\begin{equation}\label{eq:global_ob}
    {}^{1}\mathbf{r}_{f}={}^{1}\mathbf{H}_{\hat{\mathbf{x}}_{{A}_k}}\tilde{\mathbf{x}}_{{A}_k}+{}^{1}\mathbf{H}_{\hat{\mathbf{x}}_{{N}_k}}\tilde{\mathbf{x}}_{{N}_k}+{}^{1}\mathbf{H}_{^{\mathbf{kf}_1}\hat{\mathbf{p}}_{f_1}} {}^{\mathbf{kf}_1}\tilde{\mathbf{p}}_{f_1}+{}^{1}\mathbf{n}_f.
\end{equation}
The subscript ${A}$ and ${N}$ represent the active part and the nuisance part of the state vector respectively (c.f. (\ref{eq:xaxn})).

Besides, the 3D landmark ${}^{\mathbf{kf}_1}\mathbf{p}_{f_1}$ could also be directly reprojected into the anchored keyframe $\mathbf{kf}_1$ by a projection function  $^{2}\mathbf{g}$:
\begin{equation}
    \label{eq:project1}
    {}^{2}\mathbf{z}_{f}={}^{2}\mathbf{g}({}^{\mathbf{kf}_1}\mathbf{p}_{f_1})+{}^{2}\mathbf{n}_f,
\end{equation}
\begin{equation}
    \label{eq:project1_error}
    {}^{2}\mathbf{r}_{f}={}^{2}\mathbf{H}_{^{\mathbf{kf}_1}\hat{\mathbf{p}}_{f_1}} {}^{\mathbf{kf}_1}\tilde{\mathbf{p}}_{f_1}+{}^{2}\mathbf{n}_f.
\end{equation}

Furthermore, if the 3D landmark ${}^{\mathbf{kf}_1}\mathbf{p}_{f_1}$ can be observed by the other matched keyframe(s), like $\mathbf{kf}_2$ in Fig. \ref{fig2}, there is another reprojection function ${}^{3}\mathbf{g}$ which reprojects ${}^{\mathbf{kf}_1}\mathbf{p}_{f_1}$ into the $\mathbf{kf}_2$ image plane:  
\begin{equation}
    \label{eq:project2}
    {}^{3}\mathbf{z}_{f}={}^{3}\mathbf{g}({}^{G}\mathbf{T}_{\mathbf{kf}_1},{}^{G}\mathbf{T}_{\mathbf{kf}_2},{}^{\mathbf{kf}_1}\mathbf{p}_{f_1})+{}^{3}\mathbf{n}_f,
\end{equation}
\begin{equation}
    \label{eq:project2_error}
    {}^{3}\mathbf{r}_{f}={}^{3}\mathbf{H}_{\hat{\mathbf{x}}_{{N}_k}}\tilde{\mathbf{x}}_{{N}_k}+{}^{3}\mathbf{H}_{^{\mathbf{kf}_1}\hat{\mathbf{p}}_{f_1}} {}^{\mathbf{kf}_1}\tilde{\mathbf{p}}_{f_1}+{}^{3}\mathbf{n}_f.
\end{equation}
By stacking (\ref{eq:global_ob}), (\ref{eq:project1_error}) and (\ref{eq:project2_error}) we could get the 6-dimension observation for the 3D landmark ${\mathbf{p}}_{f_1}$.

In summary, whenever the current image matches with a single keyframe or multiple keyframes, for each 3D landmark ${}^{\mathbf{kf}_j}\mathbf{p}_{f_i}$ anchored in the keyframe $\mathbf{kf}_j$, we could get the observation function with the following form:
\begin{equation}
    \mathbf{r}_{f_i}= \mathbf{H}_{\hat{\mathbf{x}}_{{A}_k}}\tilde{\mathbf{x}}_{{A}_k}+\mathbf{H}_{\hat{\mathbf{x}}_{{N}_k}}\tilde{\mathbf{x}}_{{N}_k}+\mathbf{H}_{^{\mathbf{kf}_j}\hat{\mathbf{p}}_{f_i}} {}^{\mathbf{kf}_j}\tilde{\mathbf{p}}_{f_i}+\mathbf{n}_{f_i}.
\end{equation}

It is important to note that since we regard landmarks of the map as variables with uncertainties, the Jacobian matrix $\mathbf{H}_{^{\mathbf{kf}_j}\hat{\mathbf{p}}_{f_i}}$ needs to be computed. Otherwise, the landmarks would be treated as constants, like \cite{maplab}, which would make the system inconsistent. However, different from \cite{consistent}, we do not maintain map landmarks in the state vector, so we need to marginalize ${}^{\mathbf{kf}_j}\mathbf{p}_{f_i}$. For each landmark, it has at least two measurements (in the current image, and the matched one or more keyframes), so the row number of $\mathbf{H}_{^{\mathbf{kf}_j}\hat{\mathbf{p}}_{f_i}}$ is more than the dimension of the landmark. This fact guarantees that we could project $\mathbf{r}_{f_i}$ into the left null space of  $\mathbf{H}_{^{\mathbf{kf}_j}\hat{\mathbf{p}}_{f_i}}$, like (\ref{eq:local_ns}). In this way, the items related to the landmarks are eliminated while taking the uncertainties of the landmarks into consideration :
\begin{equation}\label{eq:global_error}
     \mathbf{r}_{f_i}^{*}=\mathbf{H}^{*}_{\hat{\mathbf{x}}_{{A}_k}}\tilde{\mathbf{x}}_{{A}_k}+\mathbf{H}^{*}_{\hat{\mathbf{x}}_{{N}_k}}\tilde{\mathbf{x}}_{{N}_k}+\mathbf{n}_f^{*} \triangleq \mathbf{H}^{*}_{\hat{\mathbf{x}}_{k}}\tilde{\mathbf{x}}_{k}+\mathbf{n}_f^{*}.
\end{equation}
Note that (\ref{eq:global_error}) only considers one landmark. We need to stack (\ref{eq:global_error}) for all matched landmarks to get the final observation function:
\begin{equation}\label{eq:global_final}
    \mathbf{r}^{*}_{k}=\mathbf{H}_{k}^{*}\tilde{\mathbf{x}}_{k}+\mathbf{n}^{*},
\end{equation}
where $\mathbf{r}^{*}_{k}$, $\mathbf{H}_{k}^{*}$ and $\mathbf{n}^{*}$ are the stacking of many $\mathbf{r}_{f_i}^{*}$, $\mathbf{H}^{*}_{\hat{\mathbf{x}}_{k}}$ and $\mathbf{n}_f^{*}$. (\ref{eq:global_final}) will be used to do the state update following the Schmidt-EKF as stated in the next section.

\section{State update with map information}
\label{algorithm}
In this section, we would demonstrate how to update the state with map information efficiently while maintaining the consistency. Besides, we would point out the necessity of re-linearizing the observation function.

\subsection{Global Update by Schmidt-EKF} \label{schmidt ekf}
When there are matches between the current query frame and the map keyframes, the global observations would be used to do the global update using Schmidt-EKF.
As indicated by (\ref{eq:xaxn}), the state vector could be divided into the active part and the nuisance part.
Therefore, (\ref{eq:global_final}) could be written as:
\begin{equation}\label{eq:global schmidt}
\begin{aligned}
 \mathbf{r}^{*}_{k}&=\mathbf{H}_{k}^{*}\tilde{\mathbf{x}}_{k}+\mathbf{n}^{*}\\
 &=\left[\begin{array}{cc}
     \mathbf{H}^{*}_{A_k} & \mathbf{H}^{*}_{N_k} \\
 \end{array}\right]
 \left[\begin{array}{c}
      \tilde{\mathbf{x}}_{A_k}  \\
      \tilde{\mathbf{x}}_{N_k} 
 \end{array}\right] + \mathbf{n}^{*},\\
\end{aligned}
\end{equation}
where $\mathbf{H}^{*}_{N_k}$ represents the Jacobians of the related poses of the matched map keyframes; $\mathbf{H}^{*}_{A_k}$ represents the Jacobians of the current state $\mathbf{x}_I$ and the relative transformation $\mathbf{x}_t$. Based on the expression of (\ref{eq:global schmidt}), we would analyze the limitation of standard EKF and introduce the updating technique of Schmidt-EKF.

\subsubsection{Limitation of Standard EKF}
The state and the covariance update equations for EKF are as follows:
\begin{equation}\label{eq:state_update}
    \hat{\mathbf{x}}_{k}=\hat{\mathbf{x}}_{k|k-1} + \mathbf{K}\mathbf{r}^{*}_k,
\end{equation}
\begin{equation}\label{eq:update_cov}
    \mathbf{P}_{k}=\mathbf{P}_{k|k-1}-\mathbf{K}\mathbf{H}^{*}_{k}\mathbf{P}_{k|k-1},
\end{equation}
where $\hat{\mathbf{x}}_{k|k-1}$ and $\mathbf{P}_{k|k-1}$ are the output of the propagation step (c.f. (\ref{eq:state_prop}), (\ref{eq:prop})), and
{\setlength\abovedisplayskip{1pt}
\begin{equation}\label{eq:s}
    \mathbf{S}=\mathbf{H}^{*}_{k}\mathbf{P}_{k|k-1}\mathbf{H}^{*\top}_{k}+\mathbf{R},
\end{equation}
\begin{equation}\label{eq:K}
    \mathbf{K}=\mathbf{H}^{*}_{k}\mathbf{P}_{k|k-1}\mathbf{H}^{*\top}_{k}\mathbf{S}^{-1}.
\end{equation}
}
$\mathbf{R}$ is the covariance of $\mathbf{n}^{*}$. Expanding (\ref{eq:K}),
\begin{equation}\label{eq:expand_k}
\begin{aligned}
    \mathbf{K}=\left[\begin{matrix}\mathbf{K}_{A}\\\mathbf{K}_{N}\end{matrix}\right] &= \left[\begin{matrix}\mathbf{P}_{AA_{k|k-1}}\mathbf{H}_{A_k}^{*\top} + \mathbf{P}_{AN_{k|k-1}}\mathbf{H}_{N_k}^{*\top}\\\mathbf{P}_{NA_{k|k-1}}\mathbf{H}_{A_k}^{*\top} + \mathbf{P}_{NN_{k|k-1}}\mathbf{H}_{N_k}^{*\top}\end{matrix}\right]\mathbf{S}_{k}^{-1}\\
    &=\left[\begin{matrix}\bar{\mathbf{K}}_{A}\\\bar{\mathbf{K}}_{N}\end{matrix}\right]\mathbf{S}_{k}^{-1},
    \end{aligned}
\end{equation}
and substituting (\ref{eq:expand_k}) into (\ref{eq:update_cov}), we could get:
{\setlength\abovedisplayskip{2mm}
\setlength\belowdisplayskip{-2mm}
\begin{equation}
    \label{eq:expand_update_cov}
    \begin{aligned}
      \mathbf{P}_{k}&=\mathbf{P}_{k|k-1}-\\ 
      &\left[\begin{matrix} \mathbf{K}_{A}\mathbf{S}\mathbf{K}_{A}^{\top}& \mathbf{K}_{A}\mathbf{H}_{k}^{*\top} \left[\begin{matrix} \mathbf{P}_{AN_{k|k-1}}\\\mathbf{P}_{NN_{k|k-1}}
    \end{matrix}\right]\\
    \left[\begin{matrix} \mathbf{P}_{AN_{k|k-1}}\\\mathbf{P}_{NN_{k|k-1}}\\
    \end{matrix}\right]^{\top} \mathbf{H}_{k}^{*\top} \mathbf{K}_{A}^{\top}& \mathbf{K}_{N}\mathbf{S}\mathbf{K}_{N}^{\top}\end{matrix}\right].  
    \end{aligned}
\end{equation}
}
\vspace{-0.0cm}
From (\ref{eq:expand_update_cov}), we could see that in the standard EKF update step, when we need to update the covariance, the computation is dominated by $\mathbf{K}_{N}\mathbf{S}\mathbf{K}_{N}^{\top}$, which is $\mathcal{O}(n^2)$ (suppose the dimension of $\mathbf{x}_N$ is $n$). As map information is added continuously while the robot is traveling, the size of the nuisance part $\mathbf{x}_{N}$ would grow over time. Especially for a large scene, the dimension of nuisance part can be thousands easily. In this situation, the filter-based system would suffer from the huge amount of computation and even could not run in real-time. Thanks to Schmidt-EKF, this problem could be solved elegantly.
\vspace{-0.1cm}
\subsubsection{Update by Schmidt-EKF}
As analysed before, computing $\mathbf{K}_{N}\mathbf{S}\mathbf{K}_{N}^{\top}$ is time-consuming. Therefore, in Schmidt-EKF, we just drop this part directly when we update the covariance. More specifically, we set $\mathbf{K}_N = \mathbf{0}$, so that (\ref{eq:expand_update_cov}) can be rewritten as:
{\setlength\abovedisplayskip{2mm}
\setlength\belowdisplayskip{2mm}
\begin{equation} 
    \label{eq:schmidt_update_cov}
    \begin{aligned}
      \mathbf{P}_{k}&=\mathbf{P}_{k|k-1}-\\ 
      &\left[\begin{matrix} \mathbf{K}_{A}\mathbf{S}\mathbf{K}_{A}^{\top}& \mathbf{K}_{A}\mathbf{H}_{k}^{*\top} \left[\begin{matrix} \mathbf{P}_{AN_{k|k-1}}\\\mathbf{P}_{NN_{k|k-1}}
    \end{matrix}\right]\\
    \left[\begin{matrix} \mathbf{P}_{AN_{k|k-1}}\\\mathbf{P}_{NN_{k|k-1}}\\
    \end{matrix}\right]^{\top} \mathbf{H}_{k}^{*\top} \mathbf{K}_{A}^{\top}& \mathbf{0}\end{matrix}\right],  
    \end{aligned}
\end{equation}
}
and (\ref{eq:state_update}) would be divided into two parts:
{\setlength\abovedisplayskip{2mm}
\setlength\belowdisplayskip{0mm}
\begin{equation}
\begin{aligned}
 \hat{\mathbf{x}}_{A_k}&=\hat{\mathbf{x}}_{A_{k|k-1}} + \mathbf{K}_{A}\mathbf{r}_{k}^{*},\\
    \label{eq:update_n}
    \hat{\mathbf{x}}_{N_k}&=\hat{\mathbf{x}}_{N_{k|k-1}}.
\end{aligned}
\end{equation}
}

It can be seen that the active part update is identical to the standard EKF while the nuisance part would not be updated.
Comparing (\ref{eq:expand_update_cov}) and (\ref{eq:schmidt_update_cov}), it is easy to see that:
{\setlength\abovedisplayskip{2mm}
\setlength\belowdisplayskip{2mm}
\begin{equation}
    \mathbf{P}_{SKF}=\mathbf{P}_{EKF}+ \left[\begin{matrix}\mathbf{0}\\ \bar{\mathbf{K}}_{N}\end{matrix}\right] \mathbf{S}^{-1} \left[\begin{matrix}\mathbf{0}& \bar{\mathbf{K}}_{N}^{\top}\end{matrix}\right],
\end{equation}}
where $\mathbf{P}_{SKF}$ and $\mathbf{P}_{EKF}$ are updated covariance of Schmidt-EKF and standard EKF respectively. As $\mathbf{S}^{-1}$ is postive definite, $\left[\begin{matrix}\mathbf{0}\\ \bar{\mathbf{K}}_{N}\end{matrix}\right] \mathbf{S}^{-1} \left[\begin{matrix}\mathbf{0}& \bar{\mathbf{K}}_{N}^{\top}\end{matrix}\right]$ must be positive semi-definite. Therefore, $\mathbf{P}_{SKF}-\mathbf{P}_{EKF}>=\mathbf{0}$, and it is not overconfident about the estimated state, i.e. maintains the system consistency. Besides, different from \cite{getout}, the cross-covariance of active state and map information is considered (c.f. (\ref{eq:schmidt_update_cov})) whose computation complexity is $\mathcal{O}(n)$.
\vspace{-0cm}
Therefore, for the state update with global information, we just need to linearize the global observation function to the form of (\ref{eq:global schmidt}), and substitute (\ref{eq:global schmidt}) into (\ref{eq:schmidt_update_cov})-(\ref{eq:update_n}).

\subsection{Re-linearization}\label{relinear}


In the update step, as our observation function is nonlinear, we need to linearize it at the newest estimated point:

\begin{equation}
    \label{eq:linear_ob}
    \begin{aligned}
    \mathbf{z}&=\mathbf{h}(\mathbf{x})+\mathbf{n}\\
    &=\mathbf{h}(\hat{\mathbf{x}})+ \mathbf{H}_{\hat{\mathbf{x}}}(\mathbf{x}-\hat{\mathbf{x}})+o(\mathbf{x}^2)+\mathbf{n}\\
    &\approx \mathbf{h}(\hat{\mathbf{x}})+ \mathbf{H}_{\hat{\mathbf{x}}}(\mathbf{x}-\hat{\mathbf{x}})+\mathbf{n},
    \end{aligned}
\end{equation}
where $\mathbf{H}_{\hat{\mathbf{x}}}$ is the Jacobians of the observation function with respect to $\mathbf{x}$ at the estimated point $\hat{\mathbf{x}}$. This approximation is only effective when the estimated value (linearization point) is close to the true value. When the error between the estimated value and the true value is big, the precision of this approximation is not satisfactory.

Revisited our global observation function, i.e. (\ref{eq:global_observation_function}), the landmark in the matched keyframe reference system is reprojected into the current image plane through a big ``loop'' (c.f. Fig. \ref{fig2}): The observation of the landmark $^{\mathbf{kf}_1}\mathbf{p}_{f_1}$  in current image plane (black dot in \ref{fig2}) is derived from $^{G}\mathbf{T}_{\mathbf{kf}_1}$,$^{G}\mathbf{T}_{L}$ and $^{L}\mathbf{T}_{\mathbf{C}_k}$. Therefore the error of related variables would be propagated to the final estimated measurement, which could be very big. Especially, if there is a long time lack of matching information, i.e. the system does not have global information to constrain the odometry for a long time, when the next global observation occurs, the drift of the odometry and the estimation error of $^{G}\mathbf{T}_{L}$ can be big. If we still directly linearize the observation function, the precision of the first-order approximation would be poor.



A simple idea to reduce the first-order approximation error is performing the Taylor expansion at a more accurate point. To be specific, when we get a match between a map keyframe and the current image, we could compute a relatively accurate pose ${}^{\mathbf{kf}}\mathbf{T}_{\mathbf{C}_k}$ by solving EPnP as mentioned in Sec. \ref{global model}. Based on this more accurate ${}^{\mathbf{kf}}\mathbf{T}_{\mathbf{C}_k}$, we could recompute $^{G}\mathbf{T}_{L}$ or $^{L}\mathbf{T}_{\mathbf{C}_k}$. The recomputed value would therefore be more accordant with the observation function (\ref{eq:global_observation_function}). Then, we utilize this new value as the linearization point to compute the Jacobians of the observation function.

In our implementation, when the average of the matched features' re-projection error is bigger than a threshold, the re-linearization mechanism would be triggered. We would re-compute the linearization point of $^{G}\mathbf{T}_{L}$ or $^{L}\mathbf{T}_{\mathbf{C}_k}$ from the results of EPnP. After that, $\mathbf{H}_{\hat{\mathbf{x}}}$ in (\ref{eq:linear_ob}) is computed with the re-computed linearization point.


\section{experimental results}
\label{experiemt}
\begin{figure}[t!]
\vspace{0.2cm}
    \centering
    \setlength{\abovecaptionskip}{0cm}
    \includegraphics[width=0.8\linewidth]{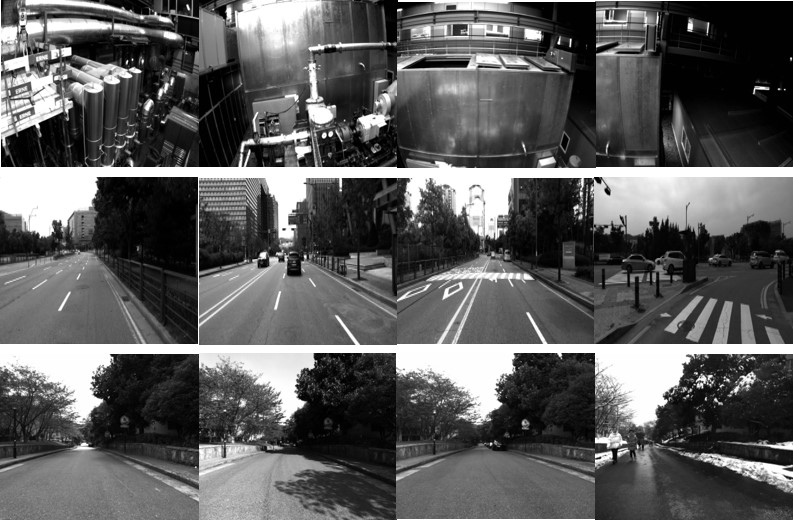}
     \caption{The images sampled from three data sets. The first row is the images from EuRoC \cite{euroc} MH02-05; The second row is the images from Kaist \cite{kaist} urban38-39; The third row is the images from YQ YQ1-4.}
     \label{fig:dataset}
    \vspace{-0.7cm}
\end{figure}

In this section, we would first do some simulation experiments to demonstrate the effectiveness of our proposed framework, and then validate our approach on EuRoC \cite{euroc}, Kaist \cite{kaist}, and our own data set YQ. These three data sets cover the platforms of UAV, car and UGV. Images sampled from these three data sets are shown in Fig. \ref{fig:dataset}.

\subsection{Simulation}\label{sec:simulation}
\subsubsection{Map Data Generation}
We make some adaptations to the simulator of Open-VINS \cite{openvins} to make it suitable for localization. The simulator needs to be fed a ground truth trajectory so that the system could generate image features and IMU information. Here, we feed the ground truth trajectory of our own data set (denoted as YQ3, see Fig. \ref{fig:sim_traj}). It has a length of 1.282km. The map keyframes' poses come from another trajectory of our own data set YQ1, which has the similar running path as YQ3. For the details of YQ data set, please refer to Sec. \ref{sec:YQ}.

To test the consistency of our proposed system, we artificially add the noise to the ground truth of YQ1 to simulate that the map is not perfect. To be specific, the position of the ground truth is perturbed with the Gaussian white noise $\mathbf{n}_p \sim \mathcal{N}(\mathbf{0},0.01\mathbf{I}_{3\times3})$, and the orientation is perturbed with $\mathbf{n}_o \sim \mathcal{N}(\mathbf{0},0.00025\mathbf{I}_{3\times3})$. After the perturbation, the root-mean-squared error (RMSE) of the trajectory of YQ1 is 0.179m.

For the map matching information, we first randomly generate 3D landmarks, then we utilize the global observation function (\ref{eq:global_observation_function}) to reproject 3D landmarks into the map keyframes and the current frame so that the 2D-2D matching features are obtained, after which we add Gaussian white noise to the 2D features. Finally, for each 3D landmark, we utilize the noisy 2D features and map keyframes' perturbed poses to triangulate its 3D position (estimated 3D landmark).

\subsubsection{Results and Analysis}
In our simulation, there are four settings: the original Open-VINS\cite{openvins}; our method (CS-MSCKF) with single matching frame (SM); our method (CS-MSCKF) with multiple matching frames (MM); our method that treats map poses and landmarks as constants (mapconst).

\begin{figure}
\vspace{2mm}
    \centering

    \subfigure[trajectories on x-y plane]{
    \begin{minipage}{1\linewidth}\centering
    \includegraphics[width=0.8\textwidth]{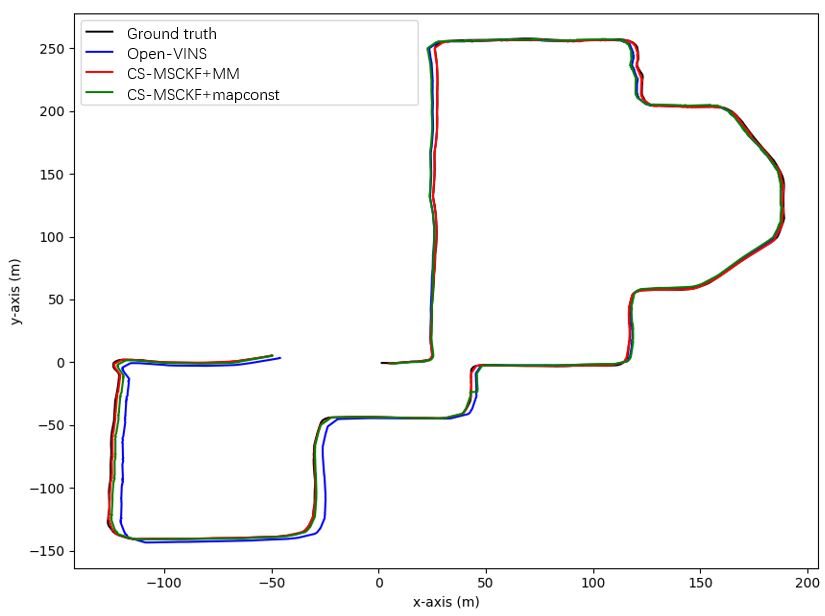}
    \end{minipage}
    }\vspace{-0.1cm}
    \subfigure[trajectories along z axis v.s. timestamps]{
    \begin{minipage}{1\linewidth}\centering
    \includegraphics[width=0.8\textwidth]{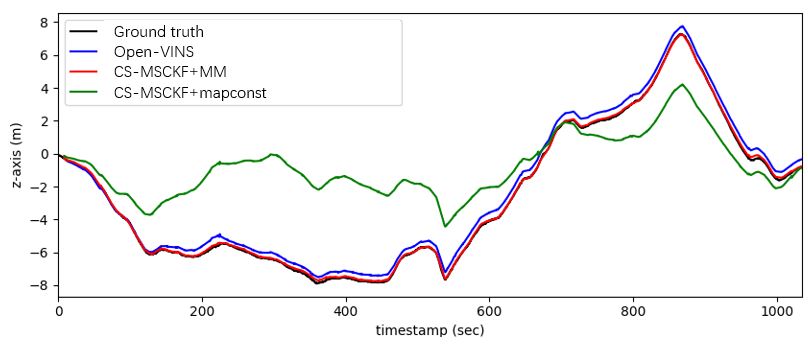}
    \end{minipage}
    }
    \setlength{\abovecaptionskip}{-0.3cm}
     \caption{The trajectory comparison of different methods}
     \vspace{-0.2cm}
     \label{fig:sim_traj}
\end{figure}

Fig. \ref{fig:sim_traj} demonstrates the trajectories derived from different situations. The blue one is the trajectory from the Open-VINS odometry. We could see that there is an apparent drift in the latter part of the trajectory, whereas our localization result (red one for multiple matching frames) fit nicely with the ground truth (black one). The green one is the result from the algorithm treating the map as perfect. The reason why the green one fit badly with the ground truth is that our map information is not absolutely accurate, and if we do not consider the uncertainty of the map, the estimator would highly rely on the inaccurate map information and therefore leading to a bad result. 

Table \ref{tab:rmse_sim} lists the RMSE of the four settings. Noting that the initial pose of odometry is set as the initial pose of the ground truth, the trajectory of the odometry is already in the global coordinate system, i.e. the initial poses of the odometry and the ground truth are naturally aligned, so that the RMSEs between the estimated trajectories and the ground truth can be computed directly without the need of alignment.

\begin{table}[!t]
\setlength{\abovecaptionskip}{0cm}
\caption{The RMSE/m of different methods}
\label{tab:rmse_sim}
\setlength{\tabcolsep}{11pt}
\begin{tabular}{c|c|c|c}
\hline
\makecell[c]{Open-VINS\\\cite{openvins}}& 
\makecell[c]{CS-MSCKF\\+SM} & 
\makecell[c]{CS-MSCKF\\+MM }&
\makecell[c]{CS-MSCKF\\+mapconst}\\
\hline
3.250 & 0.399 & 0.375 & 4.062\\
\hline
\end{tabular}
\vspace{-0.6cm}
\end{table}

From Table \ref{tab:rmse_sim}, we could find that CS-MSCKF+MM is better than CS-MSCKF+SM. It is in accordance with our intuition, as multiple frames could provide more information. Besides, as is shown in Table \ref{tab:rmse_sim}, when the map is treated as perfect (CS-MSCKF+mapconst), the performance of the algorithm is very bad, which demonstrates the necessity of taking the map's covariance into consideration.

\begin{figure}[!t]
\vspace{2mm}
\centering
\setlength{\abovecaptionskip}{0cm}
\subfigure[IMU postion (${}^{L}\mathbf{p}_{I_k}$) error with 3-$\sigma$ bounds towards CS-MSCKF+MM]{
\includegraphics[width=0.21\textwidth]{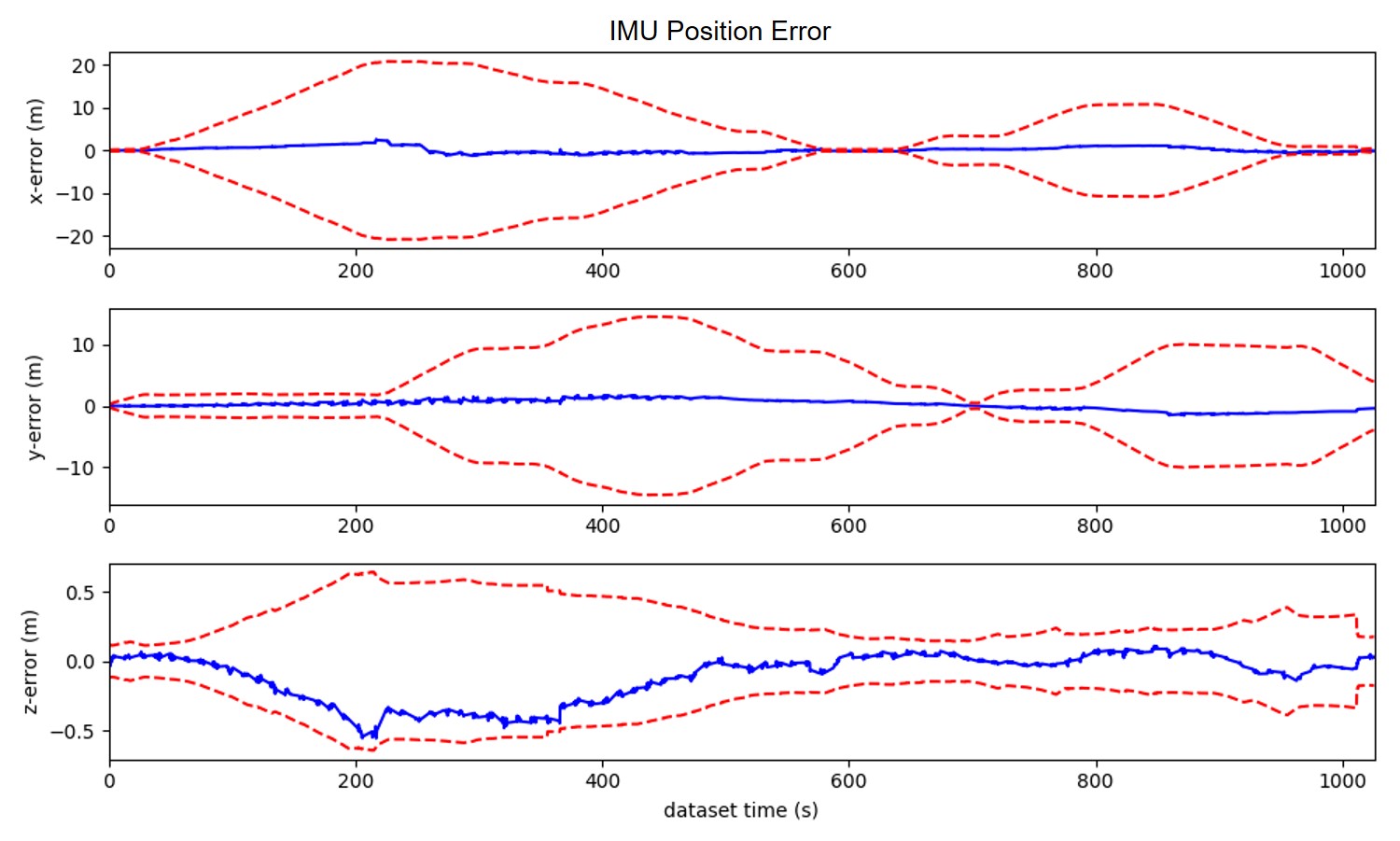}
}
\quad
\subfigure[Translation of Relative transformation ($\mathbf{x}_t$) error with 3-$\sigma$ bounds towards CS-MSCKF+MM]{
\includegraphics[width=0.21\textwidth]{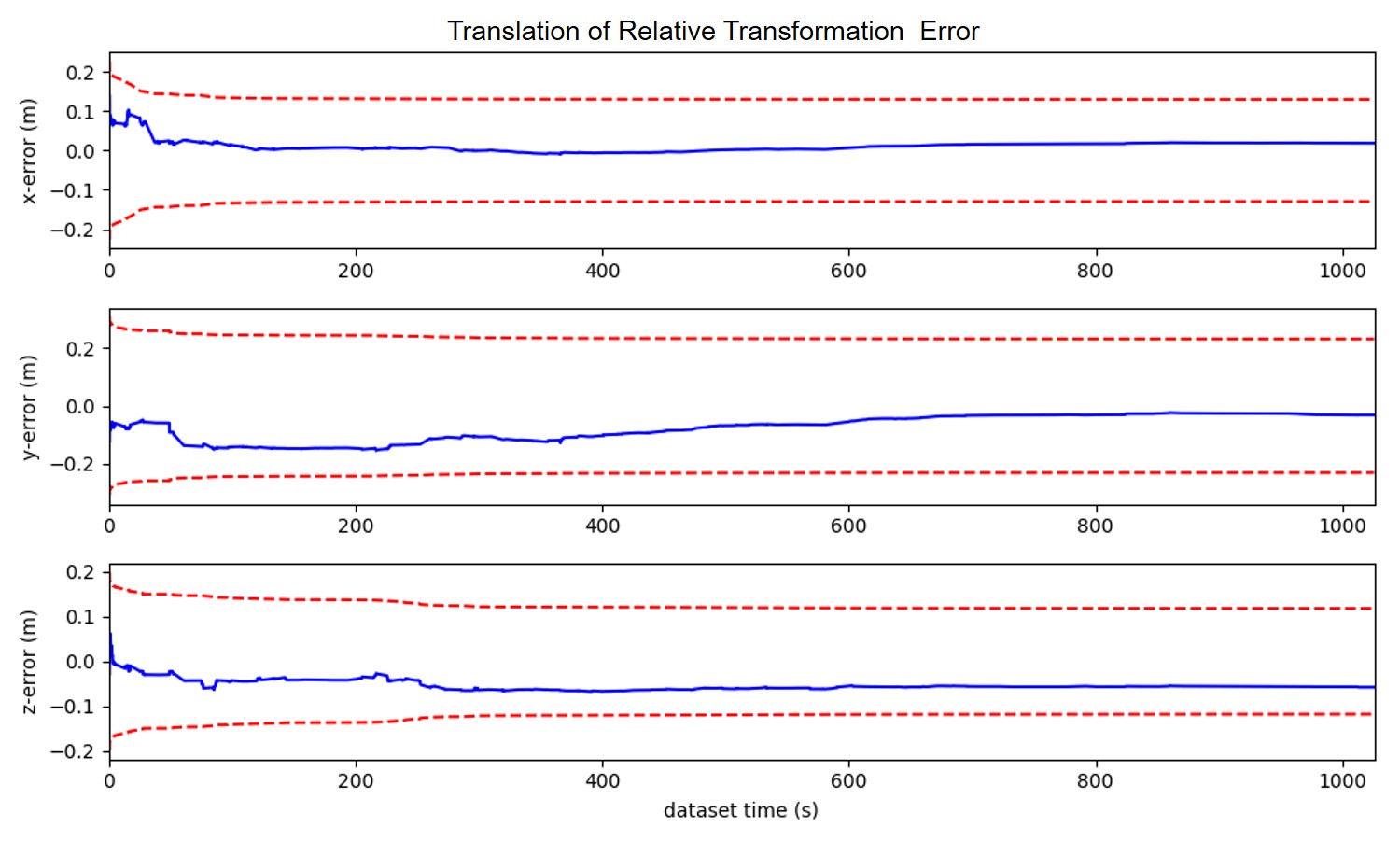}
}
\quad
\subfigure[IMU postion (${}^{L}\mathbf{p}_{I_k}$) error with 3-$\sigma$ bounds towards CS-MSCKF+mapconst]{
\includegraphics[width=0.21\textwidth]{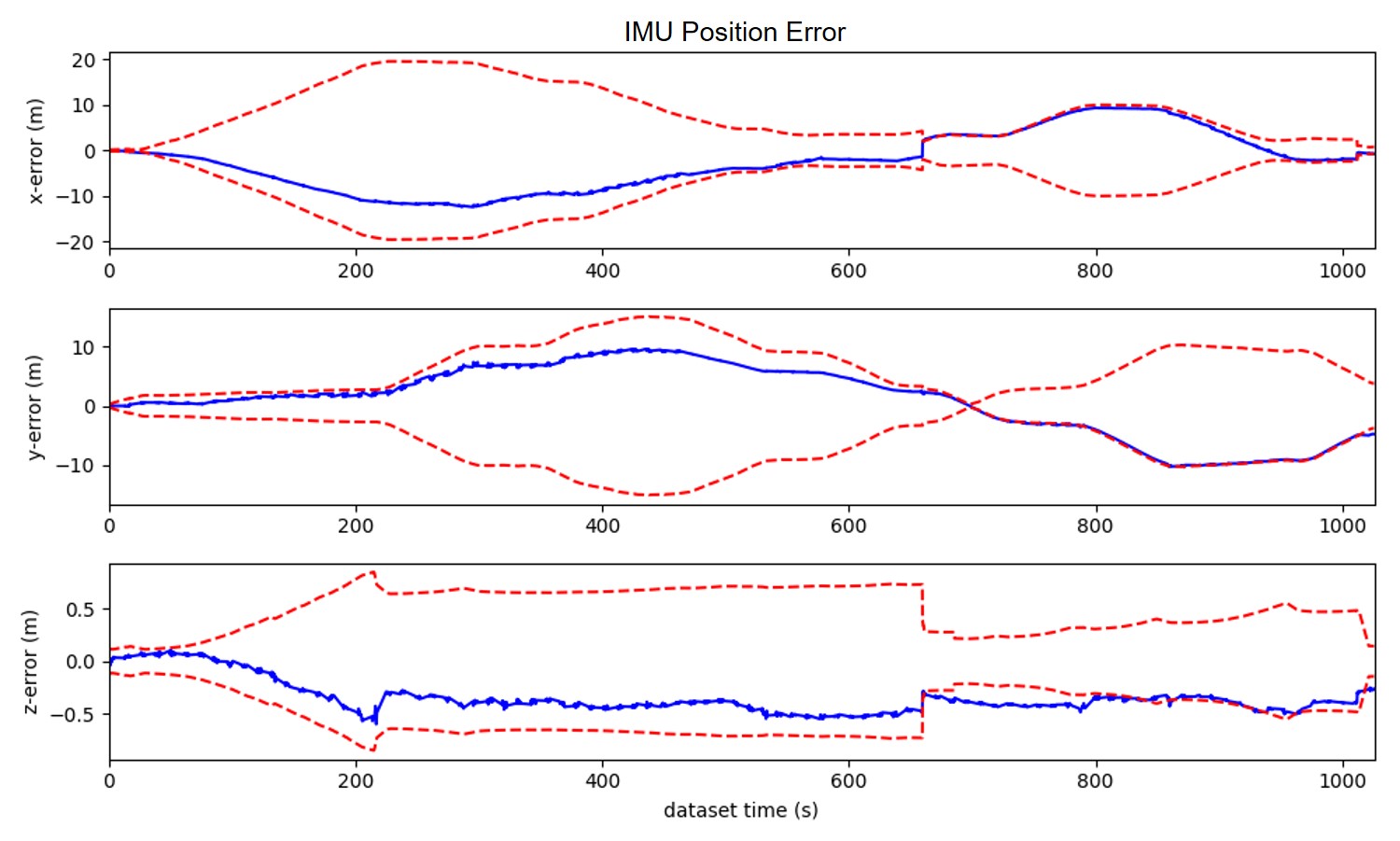}
}
\quad
\subfigure[Translation of Relative transformation ($\mathbf{x}_t$) error with 3-$\sigma$ towards CS-MSCKF+mapconst]{
\includegraphics[width=0.21\textwidth]{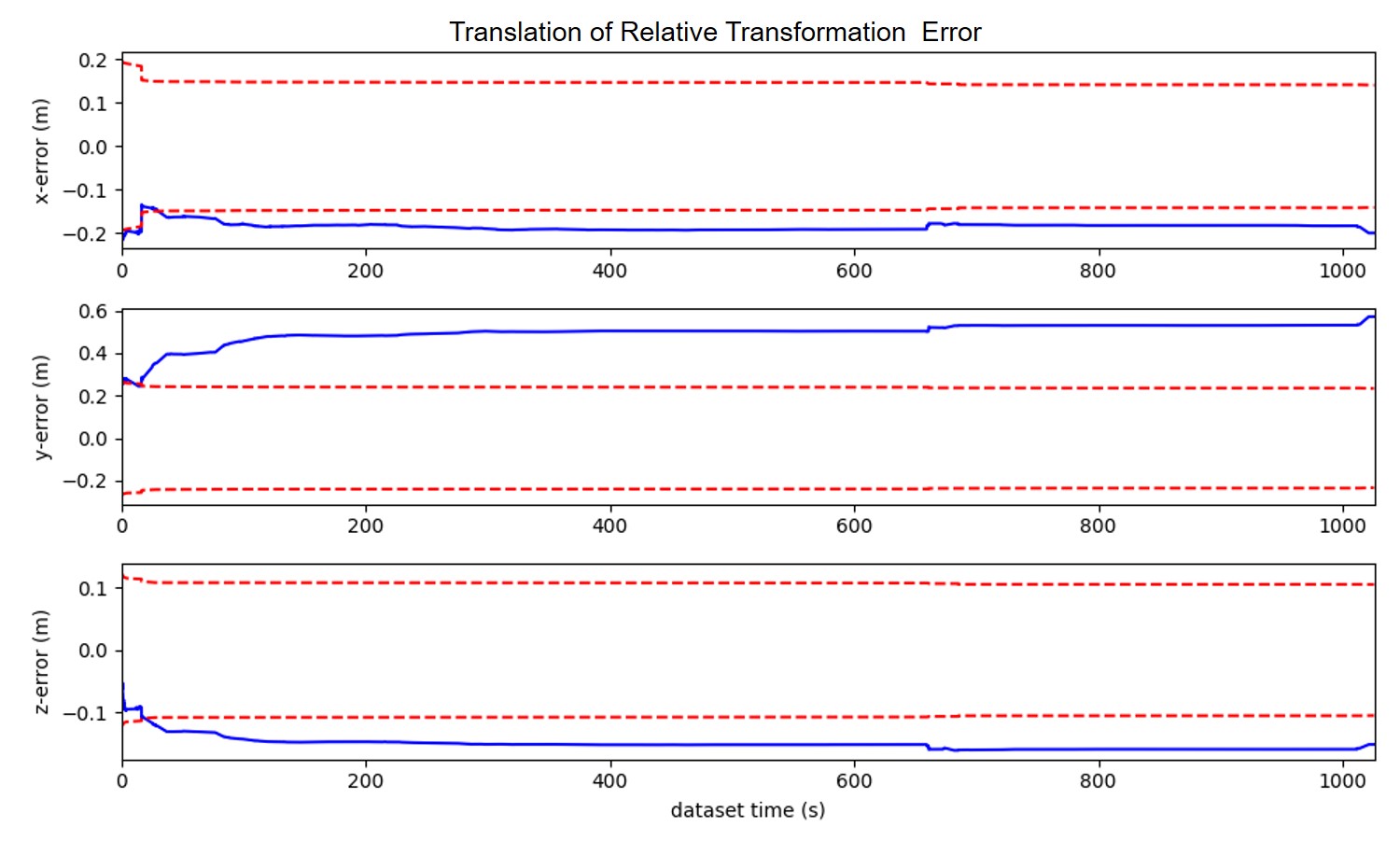}
}
\caption{The error and 3-$\sigma$ bounds towards CS-MSCKF+MM/mapconst}
\label{fig:multi_3sigma}
\vspace{-0.6cm}
\end{figure}

To demonstrate the consistency of our proposed method, we draw the error of estimation with 3-$\sigma$ bounds (c.f. Fig. \ref{fig:multi_3sigma}). The figure is corresponding to the results of CS-MSCKF+MM ((a),(b)) and CS-MSCKF+mapconst ((c),(d)). From the figure, we could conclude that our proposed algorithm has good consistency while the algorithm that treats the map information as perfect has poor consistency.

\subsection{Experiments on Real World Data Sets}
\label{sec:experiments on datasets}
For real world data sets, the matching procedure is conducted as follows: We first utilize R2D2\cite{r2d2} to extract new features on the current query frame and match with features in map keyframes. When there are enough matches, 3D-2D pairs (3D from map landmarks and 2D from current frame features) are fed into the robust pose solver in \cite{2entity}, after which the accurate and robust 3D-2D pairs are obtained.

\subsubsection{EuRoC}
In this part, the effectiveness of our proposed algorithm is validated on a popular data set EuRoC \cite{euroc}. To be specific, we use the sequences of machine hall (MH) to do experiments. The sequence MH01 is used to build the map, and the other sequences of MH (MH02-MH05) are used to do localization. The map contains keyframes and 3D landmarks, where 3D landmarks are triangulated and refined across multiple adjacent map images and the keyframes' poses are given by the ground truth (we assume the uncertainties of the ground truth are $1cm$ and $1^{\circ}$. We make the comparison between the two settings (CS-MSCKF + SM/MM) of our proposed methods with the benchmark from Open-VINS\cite{openvins} and VINS-Fusion\cite{vinsmono,vinsgps,vinslocal,vinscalib}. 

\begin{figure}[!t]
    \vspace{2mm}
    \centering
    \setlength{\abovecaptionskip}{-0.1cm}
    \includegraphics[width=0.45\textwidth]{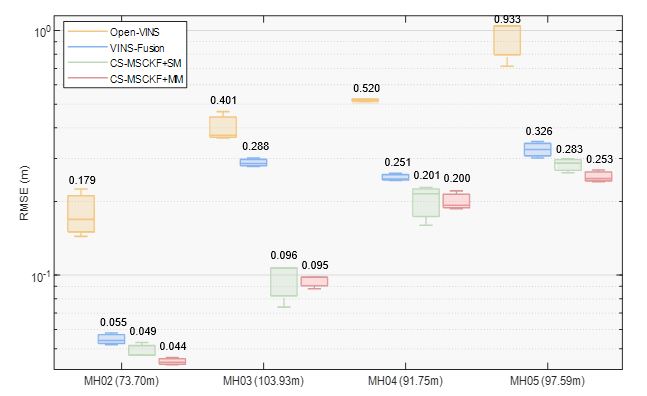}
     \caption{The RMSE/m of different methods on EuRoC\cite{euroc}}
     \label{fig:euroc_bar}
     \vspace{-0.4cm}
\end{figure}
\begin{figure}[t!]
    \centering
    \setlength{\abovecaptionskip}{-0.2cm}
    \includegraphics[width=0.45\textwidth]{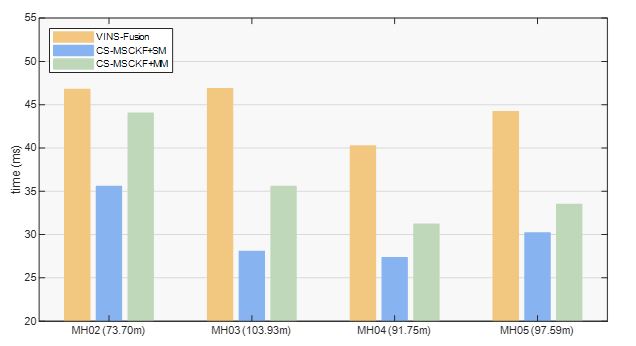}
     \caption{Time consumption (ms) of different methods on EuRoC\cite{euroc}}
     \label{fig:euroc_time}
     \vspace{-0.4cm}
\end{figure}

Open-VINS is a pure odometry, and the comparison with it could demonstrate that global localization could constrain the odometry's drift error effectively. For VINS-Fusion, its global localization mode is used. To be specific, we firstly use its SLAM mode to generate a pose-graph (map information) based on MH01, and we modified the poses of the pose-graph node as the ground truth to make the comparison fair. As the localization needs to be performed in real-time, i.e. the estimated pose in time step $k$ has to be computed in time step $k$, not the result of loop closure or global optimization thereafter, we record the global localization results from the combination of the odometry $^{L}\mathbf{T}_{k}$ (without loop closure or global optimization) and the estimated $^{G}\mathbf{T}_{L}$. The comparison of experimental results is plotted by box-plot in Fig. \ref{fig:euroc_bar}, where the RMSE of Open-VINS is computed based on the alignment of the initial pose and the rest are computed without alignments. In Fig. \ref{fig:euroc_bar}, values along $Y$-axis are projected into log space, and their specific values are also given. For the localization methods, their processing times are also given in Fig. \ref{fig:euroc_time}. All the results are the average of three runs.




From Fig. \ref{fig:euroc_bar} and Fig. \ref{fig:euroc_time}, we could find that our proposed algorithms perform better than the VINS-Fusion in terms of both accuracy and efficiency. This is because our framework take the uncertainty of the map information into consideration. Besides, the effective and robust matching algorithm provides us better matching information.  

\subsubsection{Kaist}
\begin{figure}[t!]
\vspace{2mm}
\centering
    \includegraphics[width=0.9\linewidth]{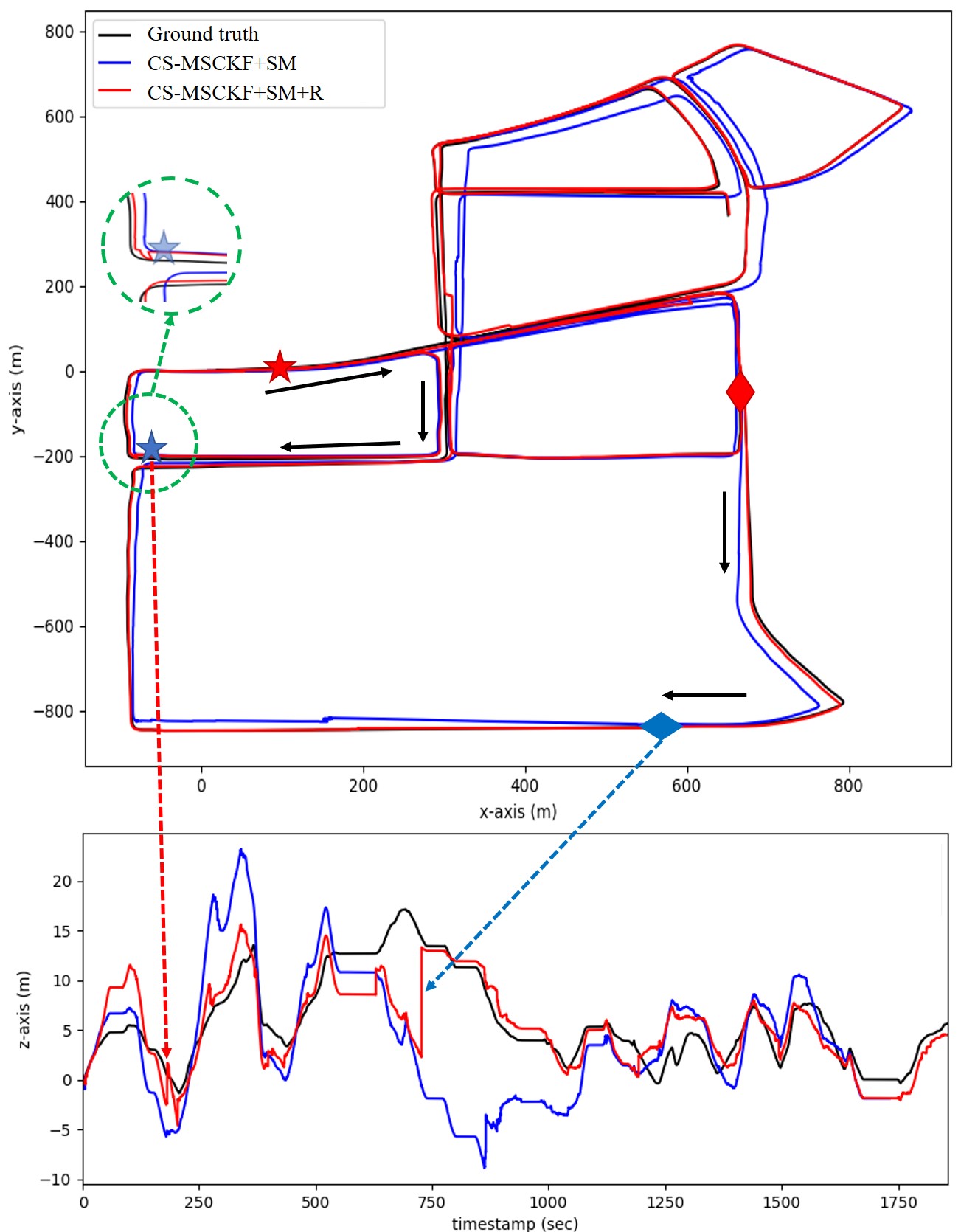}
     \caption{The trajectories derived from CS-MSCKF+SM and CS-MSCKF+SM+R.Top: trajectories on the $x-y$ plane; Bottom: trajectories along the $z$ axis v.s. time stamp; The details of the green dotted circle area is also given; The red and blue markers (stars and diamonds) represent the time when two adjacent matches are occurred (the matches at the red markers occur before the matches at the blue markers).}
     \label{fig:kaist_traj}
     \vspace{-0.6cm}
\end{figure}

To show the advantage of our proposed re-linearization mechanism, we do experiments on Kaist data set \cite{kaist}. This data set was recorded in urban environments where the majority of the scene can be taken up by other moving vehicles, so this is a challenging data set. In the data set, there are two sequences Urban38 and Urban39 with high similarity. We employ Urban38 to build the map and localization algorithms are tested on Urban39 (the length of the trajectory is 10.67km). The covariance of each map keyframe's pose is given as $0.1\mathbf{I}_{6\times6}$ and the re-linearization threshold is set as 20 pixels. The comparison of different algorithms are given in Table \ref{tab:rmse_kaist}, and the trajectories derived from CS-MSCKF+SM and CS-MSCKF+SM+R are plotted in Fig. \ref{fig:kaist_traj}.

\begin{table}[!t]
\setlength{\abovecaptionskip}{0cm}
\setlength{\belowcaptionskip}{0cm}
\caption{The RMSE/m of different methods on Kaist\cite{kaist}}
\label{tab:rmse_kaist}
\centering
\setlength{\tabcolsep}{3pt}
\begin{tabular}{c|c|c|c|c|c}
\hline
\makecell[c]{Open-\\VINS\\\cite{openvins}}&\makecell[c]{VINS-\\Fusion\\\cite{vinsmono,vinsgps}\\\cite{vinslocal,vinscalib}}& 
\makecell[c]{CS-\\MSCKF\\+SM} & 
\makecell[c]{CS-\\MSCKF\\ +SM +R} &
\makecell{CS-\\MSCKF\\ +MM }&
\makecell[c]{CS-\\MSCKF \\+MM +R} \\
\hline
34.303
  &-& 25.541 
 & 6.927
 &6.726
&\textbf{5.812}
 \\
\hline
\end{tabular}
\vspace{-0.6cm}
\end{table}

In this challenging scene, the VINS-Fusion failed to perform global localization due to the significant drifts of the odometry. From the results, we could find that for the case of single matching frame, there is a tremendous promotion in accuracy when we employ the re-linearization mechanism. Fig. \ref{fig:kaist_traj} could illustrate this phenomenon to some extent.

In Fig. \ref{fig:kaist_traj}, the pair of stars represents the two contiguous matches (for example, the $i^{th}$ match in red and the ${i+1}^{th}$ match in blue), and so does the pair of diamonds. The motion direction of the car from the red mark to the blue mark is shown by black arrows in the picture. We could find that the distance between the two stars or the two diamonds is far, which means the accuracy of the localization highly relies on the performance of the odometry. Unfortunately, the odometry inevitably introduces drifts and the estimated values of the state variables are not accurate enough. These two sources of error would make the first-order Taylor approximation of the observation function not good, as analyzed in Sec. \ref{relinear}. Therefore, we introduce a re-linearization mechanism to make the first-order Taylor approximation more accurate. The effectiveness of this mechanism is shown in Fig. \ref{fig:kaist_traj}. We could find that for the CS-MSCKF+SM, the trajectory around the blue star (c.f. the area of the green dotted circle) hardly changes when there is a match from the map, i.e. since the approximated observation function is not accurate, the single matching frame can hardly correct the trajectory. On the contrary, the trajectory from CS-MSCKF+SM+R shows an obvious sign of being corrected by the single matching frame information. Besides, as shown in the bottom part of Fig. \ref{fig:kaist_traj}, the estimated $z$ value has been corrected distinctly with the help of re-linearization mechanism while the estimated value from CS-MSCKF+SM changes little (refer to the area indicated by the red and the blue dotted arrow).  

From Table \ref{tab:rmse_kaist}, algorithms with multiple matching frames also have good performance, which is according with our intuition. This could be explained by more map information provides more constrains to the current frame so that the large drift could be corrected and bounded.

\subsubsection{YQ}\label{sec:YQ}
In this part we conduct some experiments on our own data set (YQ). This data set contains four sequences, YQ1-YQ4, where YQ1-YQ3 were recorded on three separate days with different weather in summer and YQ4 was collected in winter after snowing (c.f. Fig. \ref{fig:dataset}). For this data set, we use YQ1 to build the map, and use YQ2-YQ4 to test the performance of the algorithms. The covariance of each map keyframe's pose is given as $0.1\mathbf{I}_{6\times6}$ and the re-linearization threshold is set as 20 pixels. The comparison of different algorithms is given in Table \ref{tab:rmse_YQ}. VINS-Fusion is failed to perform continuous and consistent global localization because its odometry has large drift and its re-localization mechanism is triggered multiple times. So, the experimental results of VINS-Fusion are not listed.
\begin{table}
\vspace{2mm}
\setlength{\abovecaptionskip}{0cm}
\caption{The RMSE/m of different methods on YQ}
\label{tab:rmse_YQ}
\setlength{\tabcolsep}{7pt}
\centering
\begin{tabular}{c|c|c|c}
\hline
Sequence& 
\makecell[c]{YQ2\\(1.299km)} &\makecell[c]{YQ3\\(1.282km)} &\makecell[c]{YQ4\\(0.933km)} \\
\hline
\makecell[c]{Open-VINS \cite{openvins}} & 17.066 & 26.233 & 16.613  \\

\makecell[c]{CS-MSCKF+SM}&\makecell[c]{8.489}&\makecell[c]{9.151}&\makecell[c]{5.007}\\

\makecell[c]{CS-MSCKF+SM+R}&\makecell[c]{3.653}&\makecell[c]{3.371}&\makecell[c]{4.914}\\

\makecell[c]{CS-MSCKF+MM}&\makecell[c]{4.865}&\makecell[c]{4.779}&\makecell[c]{5.252}\\

\makecell[c]{CS-MSCKF+MM+R}&\makecell[c]{\textbf{2.869}}&\makecell[c]{\textbf{2.306}}&\makecell[c]{\textbf{4.591}}\\
\hline
\end{tabular}
\vspace{-0.6cm}
\end{table}

From the results, we could find that for the case of single matching frame, introducing re-linearization mechanism could significantly improve the positioning accuracy. Combining multiple frames matching information and re-linearization mechanism, we could get the best result.

\section{conclusion}
\label{conclusion}
In this paper, we propose a consistent filter-based global localization framework, where a keyframe-based map is used to reduce storage. Schmidt-EKF is employed to handle the problem of too much computation in traditional EKF when the dimension of the state vector is too high. In this way, the computation increases linearly with the number of map keyframes. Moreover, we introduce a re-linearization mechanism to improve the accuracy of the first-order approximation of the observation function, and therefore improve the algorithm performance, especially in large scenes. From extensive experiments, we validate the effectiveness of our proposed framework.

\end{document}